# MOTION PLANNING OF AN AUTONOMOUS MOBILE ROBOT USING ARTIFICIAL NEURAL NETWORK

G. N. Tripathi[1] and V. Rihani[2]

[1]Mody Institute of Technology and Science, Lakshamangarh, Sikar, Rajasthan
gyanendra2004@gmail.com

[2]E&EC Department, PEC University of Technology, Chandigarh, India
v.rihani@yahoo.com

## ABSTRACT

*The paper presents the electronic design and motion planning of a robot based on decision making regarding its straight motion and precise turn using Artificial Neural Network (ANN). The ANN helps in learning of robot so that it performs motion autonomously. The weights calculated are implemented in microcontroller. The performance has been tested to be excellent.*

## 1. INTRODUCTION

ANN provides high speed data processing capability of learning [1]. The use of ANN in robotics for its kinematics, dynamics, and path planning and motion control has changed the definition of the robot [2].

As per Webster a Robot is:

"An *automatic device* that performs functions normally ascribed to humans or a machine in the form of a human".

Using an ANN technique, adequate human like decision making quality can be built in a robot. This requires the analysis of environment to resolve complexity of motion and ensured that the robot is mechanically characterized to perform motion, with required degree of freedom. Then electronic has to support the motor used, hence to control the motion.

## 2. PRESENT LEVEL

Mobile robot motion planning and path planning is one of the most apparent field of application of ANN. If a robot encounters an obstacle, the arm attempts to avoid the obstacle [3]. The image processing can ensure robot obstacle avoidance and path planning in a two dimensional work space of the robot [4]. A recurrent net has been used in prediction of the motion of an object in a robot path navigation system [5]. A new learning algorithm for learning of self-organizing neural networks is proposed to recognize the traffic signs for navigation of a mobile robot in an outdoor environment [6]. Image matching is used to make decision for object [7]. An adaptive neural network can accept visual signals as inputs directly from visual sensors for the spatial information [8]. ANNs can plan motion of several mobile robots ensuring collision avoidance

[9]. Fuzzy modelling of the real robot's environment using Hopfield neural network has been used [10]. ANN base motion planer can respond to changing real time situation [11]. Latency effect in a closed loop system can be reduced for motion prediction [12]. To design the vehicle controller behavioral cloning machine learning algorithm and neural network algorithms can also be used [13]. Improvements have been made using fuzzy logic based path planning algorithm and are more effective. [14]. Convergence of a Neural Network has been improved using Q-Learning (NCQL) algorithm [15]. Extended Back Propagation Algorithm predicts moving obstacles for obstacle avoidance [16]. Designing a robot and planning its motion is always a new task in a new environment.

## 3. THE ELECTRONICS

The 2D motion of designed robot is performed using two motors, one for forward drive and another for rotation of steering to turn the vehicle direction using microcontroller is shown in the figure 3.1.

PIC16f877a has been used for the design of control circuit which has computational speed, memory (8K) and on chip ADC.

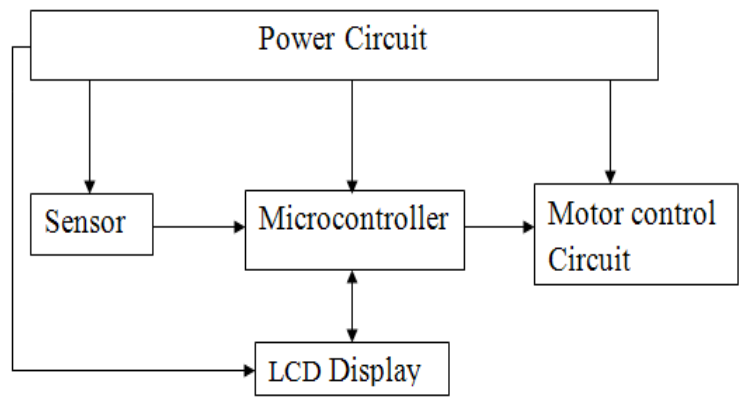

Figure 3.1 block diagram of electronic circuit

### 3.1 Sensor

Distance measuring sensor GP2Y0A21YK0F, having a PSD (position sensitive detector), IRED (infrared emitting diode) and signal processing circuit, has been used. The sensor is capable of meeting the requirements of range of distance measurement, low power consumption and package size. Output of the sensor is voltage analogous to the distance of an object.

## 4. ARTIFICIAL NEURAL NETWORK IMPLEMENTATION

Feed forward neural network is used to make decision regarding the motion of the robot.

### 4.1 The Environment

The walls of the environment are made up of Medium Density Fiber (MDF) board and wood strips. Walls are attached with each other at angle of 90 degree.

### 4.2 Application of ANN for decision making

The mobile robot designed has an Infrared sensor (a distance meter) and can scan the space in front from -90 to +90 degrees. The sensing has been made at -90, -45, 0, +45, +90 degree angles. The signal of the sensor at these angles is input to microcontroller sequentially. The ADC of microcontroller gives '8' bit digital output corresponding to the five different signals. The output of ADC is converted to binary using the threshold function.

The threshold ('th') is chosen based on the distance and corresponding ADC value sensed. Using this, ANN decides straight, left and right motion of the robot.

### 4.3 The analysis of path of environment

The top view of the path is shown here and the scanning condition for straight, left and right motion is also shown in fig.

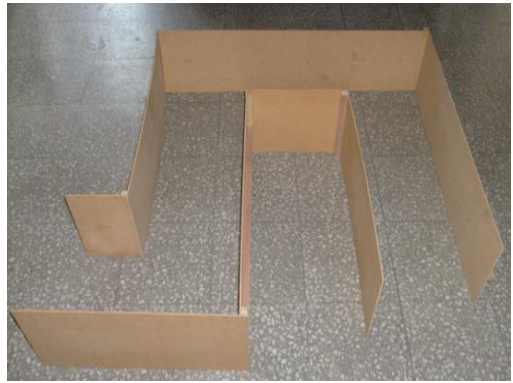

Figure-4.1(a) Real image of environment

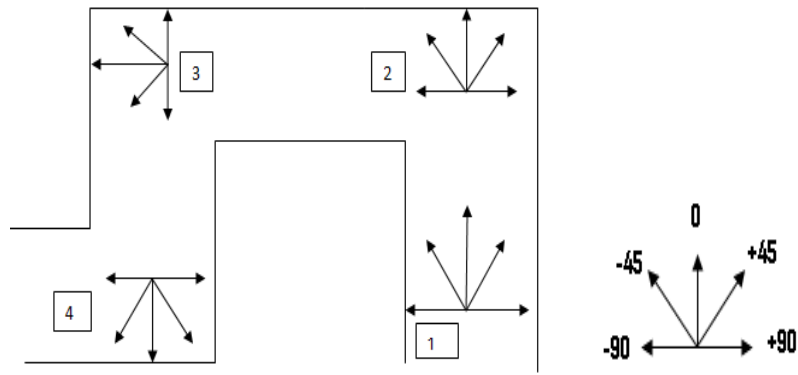

Figure-4.1(b) Top view of walls and scanning position

The point 1, 2, 3, 4 represent front area scanning point, same as shown in figure 4.1, with a changing front space.

Input to five input layer neurons is the out put of the threshold function corresponding to the ADC value sensed at angles -90, -45, 0, +45, +90. These five inputs are represented by variables $X1, X2, X3, X4, X5$ corresponding to angles -90, -45, 0, +45, +90 as in Figure 4.2.

**1.** At this position the threshold Function output for 'X1'and 'X5' will be '1' and it will represent the presence of obstacle on both left and right side of vehicle, **2.** and **3.** are the position where threshold function output for 'X3'and 'X5' will be '1' that will represent the presence of obstacle on front and right side of vehicle, **4.** at this point the threshold function Output for 'X1'and 'X3' will be '1' that will represent the presence of obstacle on front and left side of vehicle. The stopping input condition for the vehicle is given when all inputs are '1'.

Table-4.1 Input combination and corresponding outputs of ANN simulation

|    | X1 | X2 | X3 | X4 | X5 | Output |
|---|---|---|---|---|---|---|
| **1.** | 1 | 1 | 0 | 1 | 1 | straight |
|    | 1 | 0 | 0 | 1 | 1 | Straight |
|    | 1 | 1 | 0 | 0 | 1 | Straight |
|    | 1 | 0 | 0 | 0 | 1 | Straight |
|    | 0 | 0 | 0 | 0 | 0 | straight |
| **2. & 3.** | 0 | 1 | 1 | 1 | 1 | Left |
|    | 0 | 0 | 1 | 1 | 1 | Left |
|    | 0 | 1 | 1 | 0 | 1 | Left |
|    | 0 | 0 | 1 | 0 | 1 | Left |
| **4.** | 1 | 1 | 1 | 1 | 0 | Right |
|    | 1 | 1 | 1 | 0 | 0 | Right |
|    | 1 | 0 | 1 | 1 | 0 | Right |
|    | 1 | 0 | 1 | 0 | 0 | Right |
|    | 1 | 1 | 1 | 1 | 1 | No Movement |

### 4.4 Training ANN

*Easy NN-plus* has been used. It is trained using actual data and generates weights of connections for optimized condition using *Back Propagation* Algorithm and with *'tanh'* sigmoid threshold function.

### Steps used in Easy NN-plus

**Step1**. Supervised learning is used. For this both inputs and outputs are defined according to that. Data for Input to the Input Neurons and Output from Output Neurons are prepared as a first step.

**Step2.** The network is trained and the generated weight file contains the value of associations of interconnections and bias value.

### 4.5 The value of weights are given as

Weights from input layer to hidden Wij and bias Bwj.
Wij= Weight of connection from ith input layer neuron to jth hidden layer neuron.
Bwj= Biasing weight for jth hidden layer neuron.
Weights from Hidden layer to output layer Ujk and bias Buk
Ujk= weight of connection from jth hidden layer neuron to kth output layer neuron.
Buk= bias value for output layer neuron.

## 5. TESTING

Various tests are performed to carry out the measurement for achieving the desired accuracy levels. This includes the decision making.

### 5.1 Vehicle forward motion testing

This is carried out to to have practical observation that how many steps of stepper motor are needed to move a particular distance. A four '4' bit combination has been used in a loop to drive the stepper motor for multiple steps. So, number of steps = (number of loop count)* 4

Nested loops are used for driving the motor for steps more than '1024' (number of steps = 256*4). The inner loop is kept at '50' count and outer loop count was varied to observe counts it takes to move 6 inches. Due to nonlinearity of sensors, the distance from obstacle was divided in to 6 inches slot as it was observed after testing sensor that for each 6 inch slot sensor out put is linear.

Experimentally total number of steps = (number of loop count for outer loop)*(number of loop count for inner loop) *4     = 7*50*4
= 1400

## 5.2 Calibration and testing of sensor

Sensor testing is performed to get the relation of the digital output (digital output given by the controller corresponding to the sensors output voltage) to the distance from the wall. Digital value is proportional to the output voltage of sensor hence; it is proportional strength of infrared signal   reflected                                                                                         from obstacle.

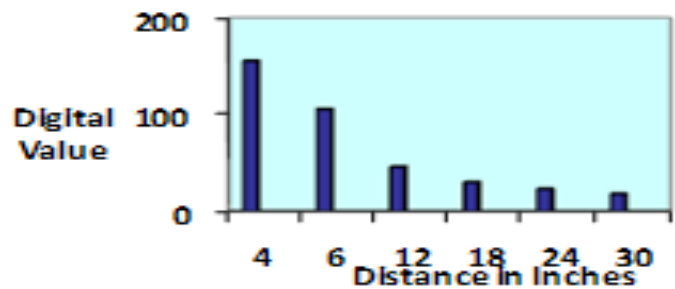

Figure-5.1 Graph chart for digital value of Sensor data to distance

## 5.3 Steering test for scanning of front area

For turning the front wheel, following steps were taken and data stored in controller. Initially it is assumed that the sensor is at '0' degree position.

1. The sensor signal, at '0' degree, is acquired and displayed to LCD
2. Sensor is turned to -45 degree, sensor signal acquired and displayed.
3. Sensor is turned to -90 degree, sensor signal acquired and displayed.
4. Sensor is turned to '0' degree position by rotating it +90 degree, sensor signal acquired and displayed again.
5. Sensor is turned to +45 degree, sensor signal acquired and displayed.
6. Sensor is turned to +90 degree, sensor signal acquired and displayed.
7. Sensor is turned to '0' degree position by rotating it -90 degree.

## 5.4 Left turn and right turn testing

The left and right turn test is done to check how many forward motion steps are required by the vehicle to complete the turning motion. It is observed that when -45 degree and + 45 degree rotation is used for left and right turn the curvature of the vehicle motion was long and the body of vehicle reaches close to the wall of environment. Therefore, to avoid collision -60 degree and +60 degree rotation is required for left turn and right turn.

The steps performed in program:
1. Rotate the wheel at -60 for left or +60 for right motion
2. Stepper Motor for forward motion is started to move forward
3. Loop count for the forward motion is displayed
4. After completion of desired turn wheel is again rotated by +60 or -60 to bring the wheel to '0' position

The required forward motion steps for complete turn are calculated as:
Total number of steps = (number of loop count for outer loop)* (number of loop count for inner loop) *4

### 5.5 Simulation test for artificial neural network

Output of ANN is used in Flowcode software. Input to the network is the output of threshold function applied to digital value of ADC, on scanning front area. The output of the network is the decision of either going straight or making a turn to left or right based on.

Threshold value 'th' = 95 has been used which is Digital value of ADC corresponding to vehicle position with respect to wall. It is the position from which vehicle needs to turn left or right, to avoid collision.

The required ANN output was achieved using '0.8' as threshold for activation function.

### 5.6 Testing of autonomous vehicle

ANN performed well to make decision for the motion of vehicle in real environment as it was noted during simulation table 4.1.

## 6. CONCLUSIONS

The designed robot is capable of moving autonomously in the environment for which it is trained. It senses the wall as obstacle and makes decision to go straight, right or left according to environment. With the use of Artificial Neural Network, self decision making by robot is performed. This robot is designed for a real time implementation of ANN for intelligent system. Use of the ANN in system improves performance.

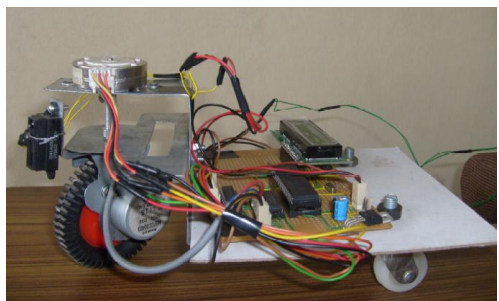

Image of Robot

**G. N. Tripathi** Received B.E. Degree from Birla Institute of Applied Sciences, Bhimatal, Uttarakhand and M.E. Degree from PEC University of Technology, Chandigarh. He is working as Assistant Professor, ECE, Mody Institute of Technology, Lakshamangarh, Rajasthan. His research interests are Microcontroller based system design, embedded design for robots and intelligent systems, Neural Network.

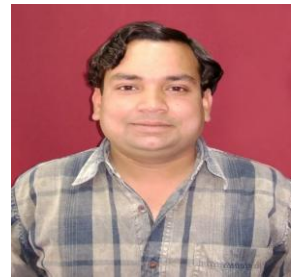

**Prof. V. Rihani** worked as faculty and HOD, E&EC department, PEC University of Technology (formerly Punjab Engineering College), Chandigarh. His research interests are Microprocessor and Controller, Computer Architecture, Neural Networks, Microwaves and Antennas.

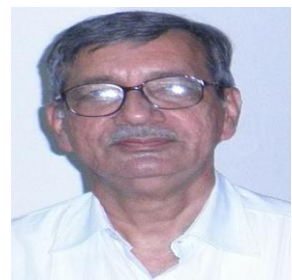